\pdfoutput=1

\documentclass[11pt]{article}

\usepackage[preprint]{acl}

\usepackage{times}
\usepackage{latexsym}

\usepackage[T1]{fontenc}

\usepackage[utf8]{inputenc}

\usepackage{microtype}

\usepackage{booktabs}
\usepackage{multirow}
\usepackage{graphicx}

\usepackage{fontawesome5}

\usepackage[normalem]{ulem}

\usepackage{enumitem}
\usepackage{caption}
\usepackage{array}
\usepackage{color}
\usepackage{soul}
\soulregister{\name}{1}

\usepackage{amsmath}
\usepackage{amssymb}
\usepackage{pifont}
\newcommand{\cmark}{\ding{51}}%
\newcommand{\xmark}{\ding{55}}%

\usepackage{cleveref}
\crefformat{section}{\textcolor{blue}{\S#2#1#3}} 
\crefformat{subsection}{\textcolor{blue}{\S#2#1#3}}
\crefformat{subsubsection}{\textcolor{blue}{\S#2#1#3}}

\usepackage[most]{tcolorbox}
\usepackage{tcolorbox}
\tcbuselibrary{theorems}

\newtcolorbox{prompt}[1]{
colback=green!5,colframe=green!35!black,fonttitle=\bfseries, title={#1}}

\usepackage{seqsplit}
\newcommand{\longtt}[1]{{\ttfamily\seqsplit{#1}}}

\newcommand\blfootnote[1]{%
  \begingroup
  \renewcommand\thefootnote{}\footnote{#1}%
  \addtocounter{footnote}{-1}%
  \endgroup
}

\newcommand{\datasetname}{\textsc{FactoidWiki}}

\title{\vspace*{-0.5in}{\normalsize \hfill EMNLP 2024 Main Conference} \\ \vspace{0.35in}Dense {\huge$\mathbb{X}$} Retrieval: What Retrieval Granularity Should We Use?}

\author{
\textbf{Tong Chen}\textsuperscript{$\clubsuit$}\thanks{$\ \ $Work was done during internship at Tencent AI Lab, Bellevue.} \,
\textbf{Hongwei Wang}\textsuperscript{$\diamondsuit$} \,
\textbf{Sihao Chen}\textsuperscript{$\heartsuit$} \,
\textbf{Wenhao Yu\textsuperscript{$\diamondsuit$}} \\
\textbf{Kaixin Ma}\textsuperscript{$\diamondsuit$} \,
\textbf{Xinran Zhao}\textsuperscript{$\spadesuit$} \,
\textbf{Hongming Zhang}\textsuperscript{$\diamondsuit$} \,
\textbf{Dong Yu}\textsuperscript{$\diamondsuit$}
\vspace{5pt} \\ 
\textsuperscript{$\clubsuit$}University of Washington \,
\textsuperscript{$\diamondsuit$}Tencent AI Lab \\
\textsuperscript{$\heartsuit$}University of Pennsylvania \,
\textsuperscript{$\spadesuit$}Carnegie Mellon University \,
}

\begin{document}
\maketitle

\begin{abstract}

Dense retrieval has become a prominent method to obtain relevant context or world knowledge in open-domain NLP tasks. When we use a learned dense retriever on a retrieval corpus at \textit{inference} time, an often-overlooked design choice is the \textit{retrieval unit} in which the corpus is indexed, e.g. document, passage, or sentence. We discover that the retrieval unit choice significantly impacts the performance of both retrieval and downstream tasks. Distinct from the typical approach of using passages or sentences, we introduce a novel retrieval unit, \textit{proposition}, for dense retrieval. Propositions are defined as atomic expressions within text, each encapsulating a distinct factoid and presented in a concise, self-contained natural language format. We conduct an empirical comparison of different retrieval granularity. Our experiments reveal that indexing a corpus by fine-grained units such as propositions significantly outperforms passage-level units in retrieval tasks. Moreover, constructing prompts with fine-grained retrieved units for retrieval-augmented language models improves the performance of downstream QA tasks given a specific computation budget.
\blfootnote{\faGithub\quad\url{https://github.com/chentong0/factoid-wiki}}
\end{abstract}

\section{Introduction}

\begin{table}[t]
\centering \small
    \vspace{0.1in}
    \begin{tabular}{p{0.18\columnwidth}|p{0.70\columnwidth}}
        \toprule
        \multicolumn{2}{l}{
        Question: What is the angle of the Tower of Pisa?} \\ \midrule
        Passage ~ Retrieval     & Prior to restoration work performed between 1990 and 2001, the tower leaned at an angle of 5.5 degrees, but \sethlcolor{pink} \hl{the tower now leans at about
        3.99 degrees}. This means the top of the Leaning Tower of Pisa is displaced horizontally 3.9 meters (12 ft 10 in) from the center. \\ \midrule
        Sentence ~ Retrieval    & Prior to restoration work performed between 1990 and 2001, the tower leaned at an angle of 5.5 degrees, but \sethlcolor{pink}\hl{the tower now leans at 
        about 
        3.99 degrees}. \\ \midrule
        Proposition ~ Retrieval & \sethlcolor{pink}\hl{The Leaning Tower of Pisa now leans at 
        about
        3.99 degrees.} \\ \bottomrule
    \end{tabular}
    
    \vspace{0.1in}

    \includegraphics[width=0.98\columnwidth]{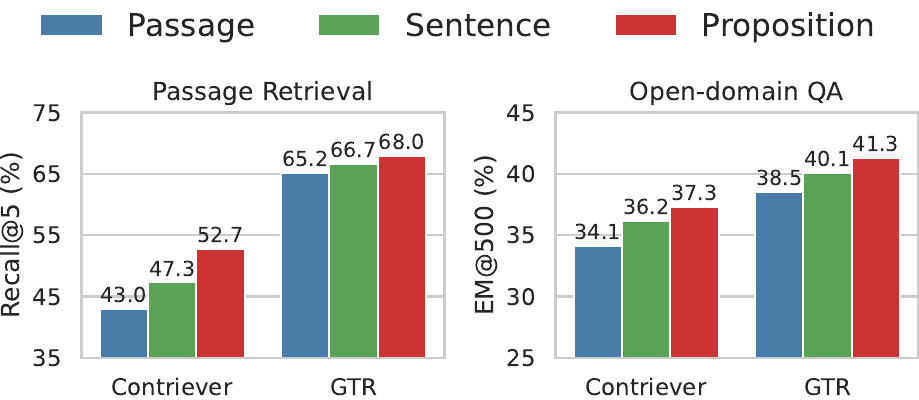}
    
    \captionof{figure}{(\textit{Top}) An example of three granularities of retrieval units of Wikipedia text when using dense retrieval. (\textit{Bottom}) We observe that retrieving by propositions yields the best retrieval performance in both passage retrieval task and downstream open-domain QA task, e.g. with Contriever \cite{izacard2022unsupervised} or GTR \cite{ni-etal-2022-large} as the backbone retriever. \sethlcolor{pink}\hl{Highlight} indicates the part that contains the answer to the question. }
    \label{fig:head}
\end{table}

\begin{figure*}[t]
    \centering
    \includegraphics[width=0.99\textwidth]{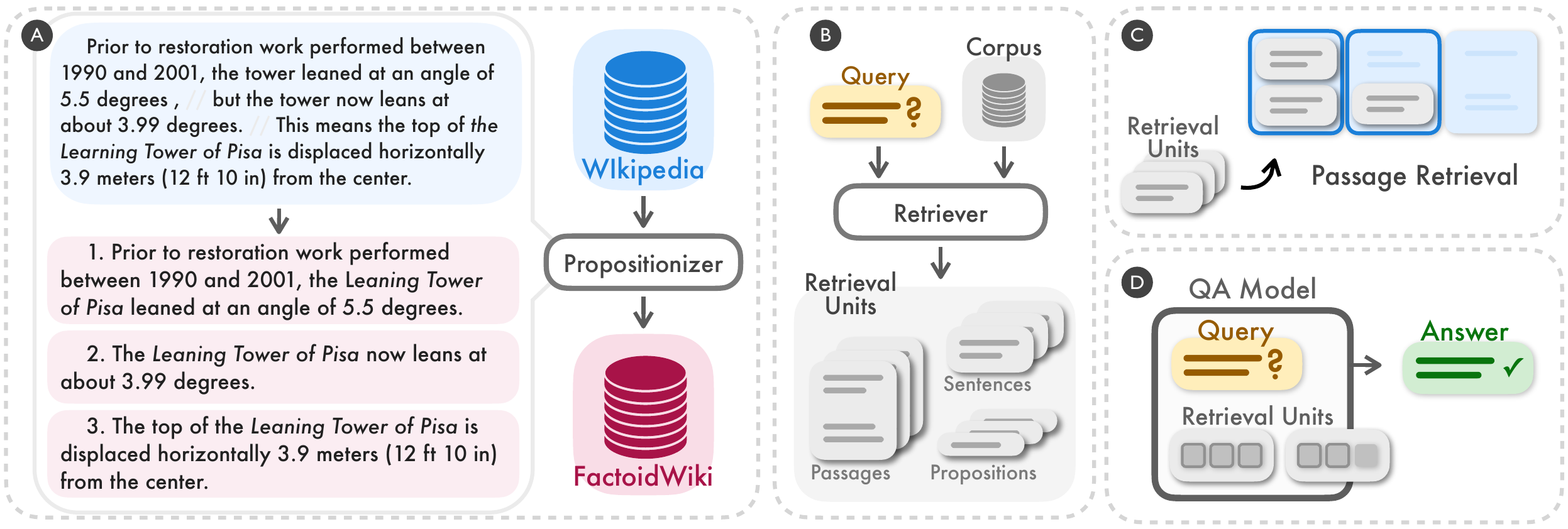}
    \vspace{-8pt}
    \caption{
    We discover that segmenting and indexing a retrieval corpus on the \textit{proposition} level can be a simple yet effective strategy to increase dense retrievers' generalization performance at inference time \textit{(A, B)}.   
    We empirically compare the retrieval and downstream open-domain QA task performance when dense retrievers work with Wikipedia indexed at the level of 100-word passages, sentences, or propositions \textit{(C, D)}. 
    }
    \label{fig:appraoch}
\end{figure*}

Dense retrievers are a popular class of techniques for accessing external information sources for open-domain NLP tasks \cite{karpukhin-etal-2020-dense}. Before we use a \textit{learned} dense retriever to retrieve from a corpus, an imperative design decision we have to make is the \textit{retrieval unit} -- i.e. the granularity at which we segment and index the retrieval corpus for inference. In practice, the choice of retrieval units, e.g. documents, fixed-length passage chunks or sentences, etc, is usually pre-determined based on how the dense retrieval model is instantiated or trained \cite{lewis2020retrieval,lee-etal-2021-learning,santhanam-etal-2022-colbertv2, ni-etal-2022-large}. 

In this paper, we investigate an overlooked research question with dense retrieval \textit{inference} -- at what retrieval granularity should we segment and index the retrieval corpus? 
We aim to investigate this question in two aspects. 

\noindent $\bullet$ First, we examine how the granularity of the index affects passage retrieval performance.

\noindent $\bullet$ Second, we investigate whether fine-grained units can replace passages in downstream QA tasks.

Based on our empirical experiments, we discover that selecting the \textit{proper} retrieval granularity at inference time can be a simple yet effective strategy for improving dense retrievers' retrieval and downstream QA performance. 
We illustrate our intuition with an example of open-domain QA in~\autoref{fig:head}. The example shows retrieved text by the same retriever at three different granularities.
The \textit{passage}, which represents a coarser retrieval unit with a longer context, is theoretically able to provide more relevant information for the question. However, a passage often includes extraneous details (e.g., restoration period and horizontal displacement in the example of~\autoref{fig:head}) that could potentially distract both the retriever and the language model in downstream tasks \cite{shi2023large,yu2023chain}.
On the other hand, \textit{sentence}-level indexing provides a finer-grained approach but does not entirely address the issue~\cite{akkalyoncu-yilmaz-etal-2019-cross,yang2020multilingual}. This is because sentences can still be complex and compounded, and they are often not self-contained, lacking necessary contextual information (e.g., in the example of \autoref{fig:head}, ``the tower'' is the coreference of ``Pisa Tower'') for judging the query-document relevance.

To address these shortcomings of typical retrieval units such as passages or sentences,
we propose using \textit{proposition} as a novel retrieval unit for dense retrieval.
Propositions are defined as atomic expressions within text, where each encapsulates a distinct factoid and is presented in a concise, self-contained natural language format. We show an example proposition in~\autoref{fig:head}. The proposition describes the information regarding the Tower of Pisa's current leaning angle in a self-contained way and precisely responds to what the question is querying. We provide a more detailed definition and description of \textit{proposition} in \cref{sec:prop}. 
To validate the efficacy of using proposition as a retrieval unit for dense retrievers inference, we first process and index an English Wikipedia dump with all documents segmented into propositions, which we refer to as \datasetname.

We conduct experiments on five different open-domain QA datasets and empirically compare the performance of four dual-encoder retrievers when Wikipedia is indexed by passages, sentences, and our proposed propositions. 
Notably, our findings indicate that proposition-based retrieval outperforms sentence and passage-based retrieval, especially in terms of generalization, as discussed in~\cref{sec:ir}. This suggests that propositions, being both compact and rich in context, enable dense retrievers to access precise information while maintaining adequate context. The average improvement over passage-based retrieval of Recall@20 is \textbf{+10.1} on unsupervised dense retrievers and \textbf{+2.7} on supervised retrievers, \textit{even though these retrievers were directly trained on passage-level retrieval.}
Furthermore, we observe a distinct advantage of proposition-based retrieval in downstream QA performance when using retrieval-augmented language models, as elaborated in ~\cref{sec:qa}. Retrieval by finer-grained units inherently provides a higher density of question-relevant information. This finding implies using finer-grained units in the prompts achieves the same performance with a shorter input length, and hence, a faster inference time.

Our main contributions are:
\begin{itemize}[leftmargin=*, itemsep=0em]
    \item {We provide a systemic study on how retrieval granularity impacts retrieval and downstream task performance. We observe that the retrieval units have a significant impact on performance.}

    \item {We introduce \datasetname, a processed English Wikipedia dump, where each page is segmented into multiple granularities: passages, sentences, and our proposed propositions. } 
    
    \item {We propose retrieval by proposition as an alternative strategy, which achieves better retrieval and QA accuracy and generalization performance (with unsupervised retriever), compared to passage or sentence as retrieval unit. } 
\end{itemize}
\section{Proposition as a Retrieval Unit}
\label{sec:prop}

The goal of our study is to understand how the granularity of a retrieval corpus influences the dense retrieval models' performance empirically. 
Aside from commonly-used retrieval units such as 100-word passages \cite{karpukhin-etal-2020-dense} or sentences, we propose using \textit{proposition} as an alternative retrieval unit choice. 
Here, propositions represent atomic expressions of meanings in text \cite{min2023factscore} with three defining principles below.  

\begin{enumerate}[leftmargin=*, itemsep=0em]
    \item {Each proposition should correspond to a distinct piece of meaning in text, where the composition of all propositions would represent the semantics of the entire text. } 
    \item {A proposition should be \textit{minimal}, i.e. it cannot be further split into separate propositions.   } 
    \item {A proposition should be \textit{contextualized and self-contained} \cite{choi-etal-2021-decontextualization}. A proposition should include all the necessary context from the text (e.g. coreference) to interpret its meaning.} 
\end{enumerate}

The use of proposition as a retrieval unit is inspired by a recent line of work \cite{min2023factscore, kamoi2023wice, chen2022propsegment, chen2023subsentence}, which finds success in representing and evaluating text semantics at the level of propositions. 
We demonstrate the concept of proposition and how a passage can be split into a set of propositions by an example on the left side of Figure \ref{fig:appraoch}. 
The passage contains three propositions, each of which corresponds to a distinct factoid about the \textit{Leaning Tower of Pisa}: the angle before the restoration, the current angle, and the horizontal displacement. 

Within each proposition, necessary context from the passage is incorporated so that the meaning of the proposition can be interpreted independently of the original text, e.g. the reference of \textit{the tower} is resolved into its full mention, \textit{the Leaning Tower of Pisa}, in the first proposition. We expect each proposition to describe exactly one atomic fact, and so our intuition is that propositions would suitably work as a retrieval unit for information-seeking questions.

\section{\datasetname: Proposition-Level Index and Retrieval for Wikipedia}
\label{sec:atomwiki}

We empirically compare passages, sentences, and propositions as retrieval units on Wikipedia, a commonly-used retrieval source for knowledge-intensive NLP tasks \cite{petroni-etal-2021-kilt}. 
To allow a fair comparison across granularities, we process an English Wikipedia dump from 2021-10-13, as used by \citet{bohnet2022attributed}. We segment each document text into three different granularities: passages, sentences, and propositions. We include the details on passage- and sentence-level segmentation of the corpus in \autoref{sec:corpus-details}.

\paragraph{Parsing Passage to Propositions.} To segment the Wikipedia pages into propositions, we finetune a text generation model, which we refer to as the \textit{Propositionizer}. The \textit{Propositionizer} takes a passage as input and generates the list of propositions within the passage. 

Following \citet{chen2023subsentence}, we train the \textit{Propositionizer} with a two-step distillation process. 
We first prompt GPT-4 \cite{OpenAI2023GPT4TR} with an instruction containing the proposition definition and 1-shot demonstration. We include the details of the prompt in \autoref{fig:prompt}. We start with a set of 42k passages and use GPT-4 to generate the seed set of paragraph-to-proposition pairs. Next, we use the seed set to finetune a Flan-T5-large model \cite{chung2022scaling}. 
We refer to the processed corpus as \datasetname. The statistics of \datasetname~are shown in \autoref{tab:stats}.

\begin{table}[t]
\centering\small
\begin{tabular}{lcc}
\toprule
            & \# units & Avg. \# words \\ \midrule
Passages     &      \ \ 41,393,528 &               58.5                            \\
Sentences    &        114,219,127 &               21.0                           \\
Propositions &        256,885,003 &               11.2                           \\ \bottomrule
\end{tabular}
\caption{Statistics of text units in the English Wikipedia.}\label{tab:stats}
\end{table}

\paragraph{Quality Analysis.}
We conduct a manual error analysis to understand the quality of propositions generated by GPT-4 and the Propositionizer. While there does not exist a fixed standard on deciding a ground truth set of propositions for a passage, we estimate the frequency of error cases where (1) a proposition is not fully supported by the passage, (2) a proposition can be further split into separate propositions, and (3) propositions are not self-contained, respectively (\autoref{tab:prop-error}). On a random sample of 50 passages, we observe that almost all propositions generated by both models are faithful, while a small portion of the propositions are not stand-alone.

\begin{table}[t]
\centering\small
\begin{tabular}{lcc}
\toprule
            & GPT-4& Propositionizer\\ \midrule
Not Faithful&        0.7\% (3/408)&               1.3\% (6/445)\\
Not Minimal&        2.9\% (12/408)&               2.0\% (9/445)\\
Not Stand-alone&        4.9\% (20/408)&               3.1\% (14/445)\\ \bottomrule
\end{tabular}
\caption{Frequency of errors occurred in the generated propositions. Most generated propositions are faithful, while a small portion of them are not stand-alone.}
\label{tab:prop-error}
\end{table}

\section{Experimental Settings}
To evaluate the impact of the three retrieval unit choices, we conduct experiments on five different open-domain QA datasets with \datasetname. 
With each dataset, we evaluate both passage retrieval and downstream QA performance when dense retrievers work with Wikipedia indexed in different granularities. 

\subsection{Open-Domain QA Datasets} 
We experiment on five different open-domain QA datasets with Wikipedia as the retrieval source: Natural Questions ~\citep[NQ,][]{kwiatkowski2019natural}, TriviaQA ~\citep[TQA,][]{joshi2017triviaqa}, Web Questions ~\citep[WebQ,][]{berant2013semantic}, SQuAD~\citep{rajpurkar2016squad}, and Entity Questions~\citep[EQ,][]{sciavolino2021simple}.

\subsection{Dense Retrieval Models}
We compare the performance of the four following supervised or unsupervised dense retriever models. Here, \textit{supervised} models refer to ones that have used human-labeled query-passage pairs as supervision during training, and vice versa.    
\vspace{-0.1in}

\begin{itemize}[leftmargin=*, itemsep=0em]
    \item \textbf{SimCSE}~\citep{gao2021simcse} is a BERT-base \cite{devlin-etal-2019-bert} encoder trained on unlabeled sentences sampled randomly from Wikipedia. SimCSE can be transferred to use as an unsupervised retriever~\citep{chen2023subsentence}.
    \item \textbf{Contriever}~\citep{izacard2022unsupervised} is an unsupervised retriever, instantiated with a BERT-base encoder. Contriever is contrastively trained by segment pairs constructed from unlabeled documents from Wikipedia and web crawl data.
    
    \item \textbf{DPR}~\citep{karpukhin-etal-2020-dense} is a dual-encoder BERT-base model fine-tuned on passage retrieval tasks directly using the question-passage pair labels from NQ, TQA, WebQ and SQuAD.

    \item \textbf{GTR}~\citep{ni-etal-2022-large} is a T5-base encoder \cite{raffel2020exploring} pretrained on online forum QA data, and fine-tuned with question-passage pair labels on MS MARCO~\citep{tri2016msmarco} and NQ datasets.
\end{itemize}

\subsection{Passage Retrieval Evaluation} 

We evaluate the retrieval performance at the passage level when the corpus is indexed at the passage, sentence, or proposition level respectively. For sentence and proposition level retrieval, we follow the setting introduced in~\citet{lee-etal-2021-phrase}, where the score of the passage is based on the maximum similarity score between the query and all sentences or propositions in a passage. 
In practice, we first retrieve a slightly larger number of text units, then map each unit to the source passage, and eventually return the top-$k$ unique passages.
We use Passage Recall@$k$ as our evaluation metric, which is defined as the percentage of questions for which the correct answer is found within the top-$k$ retrieved passages.

To further understand how different retrieved passages affect the downstream QA. We use Fusion-in-Decoder~\citep[FiD,][]{izacard-grave-2021-leveraging} model to extract answers from retrieved passages. We use a T5-large sized FiD model trained on NQ dataset in our experiments. The exact match (EM) score computes the percentage of questions for which the predicted answer exactly matches the ground truth.

\begin{table*}[t]
\centering
\resizebox{0.95\textwidth}{!}{%
\begin{tabular}{ll|cc|cc|cc|cc|cc|cc}
\toprule
\multirow{2}{*}{Retriever} & \multicolumn{1}{l|}{\multirow{2}{*}{Granularity}} & \multicolumn{2}{c|}{NQ}                    & \multicolumn{2}{c|}{TQA}                   & \multicolumn{2}{c|}{WebQ}                  & \multicolumn{2}{c|}{SQuAD}                 & \multicolumn{2}{c|}{EQ}        & \multicolumn{2}{c}{Avg.}      \\
                           & \multicolumn{1}{c|}{}                             & R@5                 & R@20                & R@5                 & R@20                & R@5                 & R@20                & R@5                 & R@20                & R@5           & R@20          & R@5           & R@20          \\ \midrule

                           

\multicolumn{14}{c}{Unsupervised Dense Retrievers} \\\midrule

SimCSE                     & Passage                                            & 28.8                & 44.3                & 44.9                & 59.4                & 39.8                & 56.0                & 29.5                & 45.5                & 28.4          & 40.3          & 34.3          & 49.1          \\
                           & Sentence                                         & 35.5                & 53.1                & 50.5                & 64.3                & 45.3                & 64.1                & 37.1                & 52.3                & 36.3          & 50.1          & 40.9          & 56.8          \\
                           & \textbf{Proposition}                             & \textbf{41.1}       & \textbf{58.9}       & \textbf{52.4}       & \textbf{66.5}       & \textbf{50.0}       & \textbf{66.8}       & \textbf{38.7}       & \textbf{53.9}       & \textbf{49.5} & \textbf{62.2} & \textbf{46.3} & \textbf{61.7} \\ \midrule
Contriever                 & Passage                                            & 42.5                & 63.8                & 58.1                & 73.7                & 37.1                & 60.6                & 40.8                & 59.8                & 36.3          & 56.3          & 43.0          & 62.8          \\
                           & Sentence                                         & 46.4                & 66.8                & 60.6                & 75.7                & 41.7                & 63.1                & 45.1                & 63.5                & 42.7          & 61.3          & 47.3          & 66.1          \\
                           & \textbf{Proposition}                             & \textbf{50.1}       & \textbf{70.0}       & \textbf{65.1}       & \textbf{77.9}       & \textbf{45.9}       & \textbf{66.8}       & \textbf{50.7}       & \textbf{67.7}       & \textbf{51.7} & \textbf{70.1} & \textbf{52.7} & \textbf{70.5} \\ \midrule

\multicolumn{14}{c}{Supervised Dense Retrievers} \\\midrule

DPR                        & Passage                                            & \ul{\textbf{66.0}} & \ul{\textbf{78.0}} & \ul{71.6}          & \ul{80.2}          & \ul{62.9}          & \ul{74.9}          & \ul{38.3}          & \ul{53.9}          & 47.5          & 60.4          & 57.3          & 69.5          \\
                           & Sentence                                         & \ul{\textbf{66.0}} & \ul{\textbf{78.0}} & \ul{\textbf{71.8}} & \ul{\textbf{80.5}} & \ul{\textbf{64.1}} & \ul{74.4}          & \ul{40.3}          & \ul{55.9}          & 53.7          & 66.0          & 59.2          & 71.0          \\ 
                           & \textbf{Proposition}                             & \ul{65.4}          & \ul{77.7}          & \ul{70.7}          & \ul{79.6}          & \ul{62.8}          & \ul{\textbf{75.1}} & \ul{\textbf{41.4}} & \ul{\textbf{57.2}} & \textbf{59.4} & \textbf{71.3} & \textbf{59.9} & \textbf{72.2} \\ \midrule
GTR                        & Passage                                            & \ul{66.3}          & \ul{78.4}          & 70.1                & 79.4                & \textbf{63.3}       & 76.5                & 54.4                & 68.1                & 71.7          & 80.5          & 65.2          & 76.6          \\
                           & Sentence                                         & \ul{66.4}          & \ul{79.4}          & 71.6                & \textbf{80.9}       & 62.2                & 76.8                & 60.9                & 73.4                & 72.5          & 81.3          & 66.7          & 78.4          \\
                           & \textbf{Proposition}                             & \ul{\textbf{66.5}} & \ul{\textbf{79.6}} & \textbf{72.2}       & \textbf{80.9}       & 63.2                & \textbf{77.4}       & \textbf{63.3}       & \textbf{75.0}       & \textbf{74.9} & \textbf{83.0} & \textbf{68.0} & \textbf{79.2} \\ \bottomrule

\end{tabular}%
}
\caption{
Passage retrieval performance (Recall@$k$ = 5, 20) on five different open-domain QA datasets when pre-trained dense retrievers work with the three different granularity from the retrieval corpus. \ul{Underline} denotes cases where the training split of the target dataset was included in the training data of the dense retriever.
}
\label{tab:ir}
\end{table*}

\subsection{Open-domain QA Evaluation on Retrieval-Augmented Language Models}

Another aspect of the choice of granularity lies in what units should be used in the prompt for retrieval-augmented language models. For large language models, retrieval-augmented generation is achieved by prepending retrieved units to user instruction and taking them as the input for language models. We aim to understand the implications of using retrieved units of different granularity within the same computational budget at inference time.
To fairly compare using different granularity in the prompts under the same computation budget, we set a token length limit for retrieved units.

For this reason, we follow an evaluation setup where the maximum number of retrieved tokens is capped at $l = 100$ or $500$, i.e. only the top $l$ tokens from passage, sentence, or proposition level retrieval are fed into the language model as input. We evaluate the percentage of questions for which the predicted answer exactly matches (EM) the ground truth. We denote our metric as EM @ $l$ tokens. 
We use LLaMA-2-7B \citep{touvron2023llama} in our evaluation. To ensure the model's output aligns with the format of each dataset, we employ in-context learning, incorporating \textbf{four-shot} demonstrations as illustrated in \autoref{fig:llama-prompt}.

\begin{figure*}[h]
    \centering
    \includegraphics[width=0.75\textwidth]{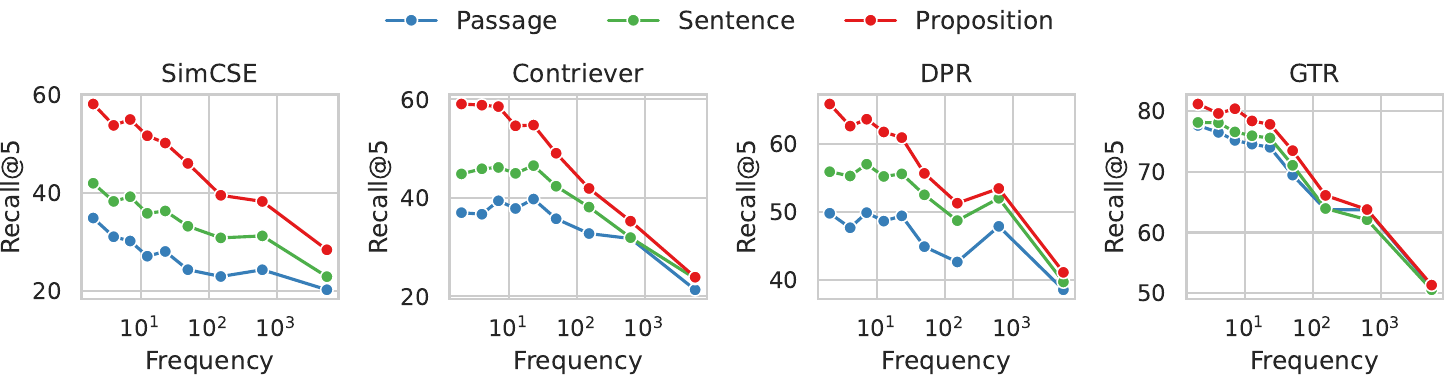}
    \vspace{-8pt}
    \caption{Document retrieval recall vs. the frequency of the target entity in each question from the \textit{Entity Questions} dataset. The frequency of each entity (i.e. smaller value $\Rightarrow$ less common entities, and vice versa) is estimated by the frequency of the entity in its top-1000 passage retrieved by BM25. On queries with less common entities, we observe that retrieving by proposition shows a larger advantage over retrieval by proposition.}
    \label{fig:entity}
\end{figure*}

\begin{table*}[t]
\centering
\resizebox{0.95\textwidth}{!}{%
\begin{tabular}{ll|cc|cc|cc|cc|cc|cc}
\toprule
\multirow{2}{*}{Retriever} & \multicolumn{1}{l|}{\multirow{2}{*}{Granularity}} & \multicolumn{2}{c}{NQ}                    & \multicolumn{2}{c}{TQA}                   & \multicolumn{2}{c}{WebQ}                  & \multicolumn{2}{c}{SQuAD}                 & \multicolumn{2}{c}{EQ}        & \multicolumn{2}{c}{Avg.}      \\ \cmidrule(l){3-14} 
                           & \multicolumn{1}{c|}{}                             & top-5& top-20& top-5& top-20& top-5& top-20& top-5& top-20& top-5& top-20& top-5& top-20\\ \midrule
\multicolumn{14}{c}{Unsupervised Dense Retrievers} \\\midrule

SimCSE                     & Passage                                            & 16.6& 23.6& 32.3& 40.8& 15.5& 19.1& 14.6& 20.7& 16.1& 20.3& 19.0& 24.9\\
                           & Sentence                                         & 20.7& 28.1& 36.0& 44.5& 18.5& 21.9& 19.6& 25.8& 19.9& 25.1& 23.0& 29.1\\
                           & \textbf{Proposition}                             & \textbf{24.5}& \textbf{33.1}& \textbf{37.5}& \textbf{46.2}& \textbf{19.7}& \textbf{23.0}& \textbf{21.4}& \textbf{27.6}& \textbf{26.8}& \textbf{32.0}& \textbf{26.0}& \textbf{32.4}\\ \midrule
Contriever                 & Passage                                            & 23.2& 35.1& 40.8& 50.8& 16.3& 22.1& 23.9& 32.7& 20.2& 27.9& 24.9& 33.7\\
                           & Sentence                                         & 26.0& 36.8& 43.4& 52.9& 18.4& 23.9& 26.7& 34.7& 23.7& 30.3& 27.6& 35.7\\
                           & \textbf{Proposition}                             & \textbf{28.9}& \textbf{39.2}& \textbf{47.2}& \textbf{55.6}& \textbf{19.5}& \textbf{25.2}& \textbf{30.8}& \textbf{37.6}& \textbf{28.8}& \textbf{35.8}& \textbf{31.1}& \textbf{38.7}\\ \midrule

\multicolumn{14}{c}{Supervised Dense Retrievers} \\\midrule

DPR                        & Passage                                            & \textbf{41.1}& \textbf{45.6}& 50.6& 57.0& 23.7& 25.5& 18.8& 25.4& 25.3& 29.7& 31.9& 36.6\\
                           & Sentence                                         & 40.3& \textbf{45.6}& \textbf{51.7}& \textbf{57.6}& 24.0& 26.9& 21.1& 27.4& 28.6& 32.9& 33.1& 38.1\\ 
                           & \textbf{Proposition}                             & 39.7& 45.2& 51.0& 56.8& \textbf{24.3}& \textbf{27.5}& \textbf{22.2}& \textbf{28.3}& \textbf{32.0}& \textbf{36.0}& \textbf{33.9}& \textbf{38.8}\\ \midrule
GTR                        & Passage                                            & 39.8& 46.1& 49.7& 55.9& 23.0& 25.9& 29.9& 35.1& 37.8& 39.6& 36.0& 40.5\\
                           & Sentence                                         & 39.4& 45.9& 51.7& 58.0& 23.2& 26.1& 35.7& 39.1& 38.0& 39.9& 37.6& 41.8\\
                           & \textbf{Proposition}                             & \textbf{40.0}& \textbf{46.9}& \textbf{52.5}& \textbf{58.4}& \textbf{24.2}& \textbf{26.5}& \textbf{37.8}& \textbf{40.4}& \textbf{39.2}& \textbf{41.0}& \textbf{38.7}& \textbf{42.6}\\ \bottomrule

\end{tabular}%
}
\caption{
Open-domain QA performance (Exact Match) using Fusion-in-Decoder model \citep{izacard-grave-2021-leveraging} to extract answer from top-5 and top-20 passages retrieved on the index of passages, sentences, and propositions.
}
\label{tab:ir-fid}
\end{table*}

\section{How Does Granularity Influence Passage Retrieval?}\label{sec:ir}

In this section, we report and discuss how indexing the corpus at various granularity influences the passage retrieval performance.
Surprisingly, despite all of the dense retrieval models being trained on only passage-level documents, all the models demonstrate on-par or superior performance when the corpus is indexed at the proposition level. 
Our results suggest that indexing the corpus at the finer-grained units improves the cross-task generalization on passage retrieval.

\subsection{Passage Retrieval Performance}\label{ssec:retrieval-results}

We report our evaluation results in \autoref{tab:ir}. We observe that retrieval by propositions outperforms retrieval by sentences or passages on most tasks for both unsupervised and supervised retrievers. 

With all dense retrievers tested, proposition-level retrieval consistently outperforms sentence and passage-level retrieval on average across the five datasets. 
With the \textit{unsupervised} retrievers, i.e. SimCSE and Contriever, we see an averaged Recall@5 improvement of $+12.0$ and $+9.3$ (35.0\% and 22.5\% relative improvement) on five datasets.

With the \textit{supervised} retrievers, proposition-level retrieval still shows an advantage on average, yet the sizes of improvements are smaller. We hypothesize that this is due to these retrievers being trained on query-passage pairs. 
For instance, with DPR, which have been trained on NQ, TQA, WebQ, and SQuAD, we observe that proposition and sentence level retrieval perform slightly worse compared to passage level on three out of the four datasets, with the exception of SQuAD. 
As shown in \autoref{tab:ir}, all supervised retrievers demonstrate comparable performance across three levels of retrieval granularity in NQ, TQA, and WebQ. 

However, on datasets that the retriever model has \textit{not} seen during training, we observe that retrieval by proposition demonstrates a clear advantage.    
For instance, most notably on SQuAD or EntityQuestions, we observe that proposition-based retrieval significantly outperforms the other two granularities. 
We see 25\% Recall@5 relative improvement on EntityQuestions with relatively weak retrievers like DPR. Furthermore, the Recall@5 of retrieval by proposition on SQuAD improved most on GTR, with 16\% relative improvements. 

\subsection{Retrieval on Finer-grained Index $\Rightarrow$ Better Cross-Task Generalization}\label{ssec:generalization}

Our results show the advantage of retrieval on proposition-level index in cross-task generalization settings. 
We observe that on SQuAD and Entity Questions, retrieval on the proposition-level index brings more performance gain over the passage-level index and sentence-level index.

To better understand where the improvements can be attributed, we conduct an additional analysis on Entity Questions. As \textit{Entity Questions} features questions targeting the properties of longer-tail entities, we study how the retrieval performance under three different granularities is affected by the \textit{occurance} of the target entity in question, i.e. whether the entity appears frequently in Wikipedia or not. 
We estimate the frequency of each entity with the following method. 
Given the surface form of an entity, we use BM25 to retrieve the top 1000 relevant passages from Wikipedia. We use the number of occurrences of the entity in its relevant passages as an estimate of its frequency.  With the 20,000 test queries, around 25\% of the target entities have an frequency value of less or equal to 3. 

\autoref{fig:entity} shows the passage retrieval performance vs. the frequency of the target entity in each question.  Across all four dense retrievers, we observe that retrieving by proposition shows a much larger advantage over retrieving by passages with questions targeting less common entities. 
As the frequency of entities increases, the performance gap decreases. 
Our findings indicate that the performance gain from retrieval by proposition can mostly be attributed to queries for long-tailed information. This echoes our observation that retrieval on proposition-level index improves the cross-task generalization performance of dense retrievers.

\subsection{Higher Passage Recall $\Rightarrow$ Higher Downstream QA Accuracy}

To further understand whether the passage retrieval on a finer-grained index achieves higher downstream QA performance, we extract the answer from the retrieved passage by a QA reader, Fusion-in-decoder. The results are shown in \autoref{tab:ir-fid}.

Retrieval by proposition-level index achieves the highest average exact match (EM) on all four retriever models. Apart from limited exceptions, the proposition-level index achieves the highest EM for most retrieval tasks and on most datasets. We observe that the trend of downstream QA performance is highly consistent with passage retrieval recall, suggesting higher passage recall implies better downstream QA performance. 

\begin{table*}[]
\centering
\resizebox{0.95\textwidth}{!}{%
\begin{tabular}{ll|cc|cc|cc|cc|cc|cc}
\toprule
\multirow{3}{*}{Retriever} & \multicolumn{1}{l|}{\multirow{3}{*}{Granularity}} & \multicolumn{2}{c}{NQ}                    & \multicolumn{2}{c}{TQA}                   & \multicolumn{2}{c}{WebQ}                  & \multicolumn{2}{c}{SQuAD}                 & \multicolumn{2}{c}{EQ}        & \multicolumn{2}{c}{Avg.}      \\ \cmidrule(l){3-14} 
                           & \multicolumn{1}{c|}{}                             & \multicolumn{2}{c|}{EM}                    & \multicolumn{2}{c|}{EM}                    & \multicolumn{2}{c|}{EM}                    & \multicolumn{2}{c|}{EM}                    & \multicolumn{2}{c|}{EM}        & \multicolumn{2}{c}{EM}        \\
                           & \multicolumn{1}{c|}{}                             & @100                & @500                & @100                & @500                & @100                & @500                & @100                & @500                & @100          & @500          & @100          & @500          \\ \midrule
\multicolumn{2}{c|}{Closed-book}& \multicolumn{2}{c|}{23.4}& \multicolumn{2}{c|}{57.4}& \multicolumn{2}{c|}{25.9}& \multicolumn{2}{c|}{13.0}& \multicolumn{2}{c|}{23.2}& \multicolumn{2}{c}{28.6}\\ \midrule
\multicolumn{14}{c}{Unsupervised Dense Retrievers}             \\ \midrule
SimCSE                     & Passage                                            & 20.5& 22.9& 49.7& 52.9& 24.5& 24.6& 13.7& 16.6& 20.7& 25.5& 25.8& 28.5\\
                           & Sentence                                         & 21.1& 24.3& \textbf{52.1}& \textbf{54.2}& \textbf{24.2}& 26.1& 17.7& 21.5& 22.9& 28.3& 27.6& 30.9\\
                           & \textbf{Proposition}                             & \textbf{22.0}& \textbf{26.0}& 51.0& 53.9& 23.5& \textbf{27.0}& \textbf{18.6}& \textbf{22.7}& \textbf{25.9}& \textbf{33.6}& \textbf{28.2}& \textbf{32.6}\\ \midrule

Contriever                 & Passage                                            & 24.5& 28.7& 54.7& 57.9& 25.7& 26.9& 17.7& 24.2& 25.6& 32.5& 29.6& 34.1\\
                           & Sentence                                         & 25.0& 30.2& 56.3& 59.2& 26.8& 29.2& 22.5& 28.1& 26.1& 34.1& 31.3& 36.2\\
                           & \textbf{Proposition}                             & \textbf{25.8}& \textbf{30.3}& \textbf{56.8}& \textbf{60.0}& \textbf{26.8}& \textbf{29.9}& \textbf{24.8}& \textbf{29.7}& \textbf{27.1}& \textbf{36.5}& \textbf{32.3}& \textbf{37.3}\\ \midrule
\multicolumn{14}{c}{Supervised Dense Retrievers}                  \\ \midrule

DPR                        & Passage                                            & 30.6& 33.7& 56.5& 60.3& 25.0& 26.8& 14.2& 18.9& 26.4& 31.6& 30.6& 34.3\\
                           & Sentence                                         & \textbf{32.5}& \textbf{34.1}& \textbf{58.3}& \textbf{61.7}& 25.4& 28.0& 17.6& 22.1& 29.8& 35.6& 32.7& 36.3\\
                           & \textbf{Proposition}                             & 31.5& 33.8& 57.6& 60.6& \textbf{27.1}& \textbf{28.2}& \textbf{18.2}& \textbf{22.6}& \textbf{32.9}& \textbf{39.7}& \textbf{33.5}& \textbf{37.0}\\ \midrule

GTR                        & Passage                                            & 30.0& 33.9& 56.9& 60.0& 24.5& 25.9& 21.5& 27.4& 42.2& 45.3& 35.0& 38.5\\
                           & Sentence                                         & 30.9& \textbf{34.0}& \textbf{58.9}& 61.9& 24.5& 27.0& 29.8& 31.7& 42.9& 45.9& 37.4& 40.1\\
                           & \textbf{Proposition}                             & \textbf{32.1}& 33.8& 58.8& \textbf{62.3}& \textbf{25.7}& \textbf{29.1}& \textbf{32.5}& \textbf{33.1}& \textbf{43.0}& \textbf{48.1}& \textbf{38.4}& \textbf{41.3}\\ \bottomrule

\end{tabular}%
}
\caption{
Open-domain QA performance (EM = Exact Match) with LLaMA-2-7B model \citep{touvron2023llama}. The context in the prompts is constructed by passage, sentence, or propositions limiting at $l=100$ or $500$ tokens. We prompt the LLaMA-2-7B model with four-shot demonstrations for each test case.}
\vspace{-5pt}
\label{tab:qa-llama}
\end{table*}

\begin{figure*}[t]
    \includegraphics[width=0.99\textwidth]{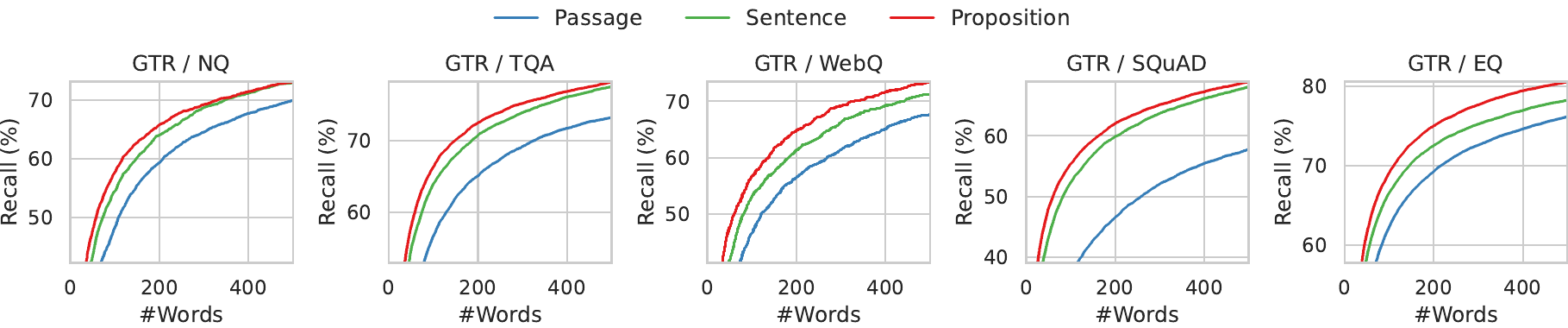}
    \vspace{-0.1in}
    \caption{Recall of the gold answer in the retrieved text limited to first $k$ words for the GTR retriever. Finer-grained retrieval has a higher recall across all numbers of words.}
    \label{fig:position}
\end{figure*}

\section{How Does Granularity Influence Retrieval-Augmented LMs?}\label{sec:qa}

In this section, we study how the choice of different granularity used in the prompts affects the retrieval-augmented generation across open-domain QA tasks. 
To fairly compare different granularity with the same computation budget, we limit the number of retrieved tokens for input to the language model at $l=100$ or $500$ tokens. 
Our results suggest that retrieval by finer-grained units enables a higher density of question-related information in the prompts, leading to better performance.

\subsection{Open-domain QA Performance}\label{ssec:odqa-results}

\autoref{tab:qa-llama} shows the evaluation results with LLaMA-2-7B as the language model. Across different retrievers, we observe higher QA performance in terms of the EM@$l$ metric on average when using propositions as the retrieval unit.

Using propositions rather than passages in the prompts, the four dense retrievers—SimCSE, ConRetriever, DPR, and GTR—improve by +4.1, +3.2, +2.7, and +2.8 in the EM@500 score. The improvements for using sentences over passages for the four retrieval models are +2.4, +2.1, +2, and +1.6, respectively. It is interesting to note that in the LLaMA-2-7B model, the QA accuracy on TQA and WebQ is not sensitive to retrieval type. The highest improvements over the closed-book setting are only +4.9 and +3.2, achieved by GTR with propositions. Nevertheless, we observe that using sentences and propositions in the prompts results in higher performance than using passages for all retrieval models on these two datasets.
The results suggest that using finer-grained units in the prompts is beneficial to retrieval-augmented generation.

\subsection{Finer-grained Granularity $\Rightarrow$ Higher Density of Question-Related Information}\label{sec:density}
Intuitively, compared to sentences or passages as retrieval units, the advantage of propositions is that the retrieved propositions have a higher density of relevant information to the query. With finer-grained retrieval units, the correct answer to the query would more likely appear in the top-$l$ retrieved words by a dense retriever.   

We illustrate this phenomenon by an analysis shown in \autoref{fig:position}. Here, we investigate the position at which the ground truth answer appears in the top-$l$ retrieved words.     
Specifically, we calculate the recall of the gold answer within the initial $l$ retrieved words with GTR working with Wikipedia indexed in three different granularities.

We show the results in \autoref{fig:position} and \ref{fig:position-all} with $l$ ranging from 0 to 500 across all five datasets. 
For a fixed word retrieval budget, proposition retrieval shows a higher success rate than sentence and passage retrieval methods. The largest improvement of proposition retrieval over passage retrieval occurs within the range of 100-200 words, which corresponds to roughly 10 propositions, 5 sentences, or 2 passages. As word count increases, the recall rate of the three granularities converges, encompassing all relevant information.

\section{Related Work}
\label{sec:related}

Recent works on dense retrievers typically adopt a dual-encoder architecture \cite{yih-etal-2011-learning, reimers-gurevych-2019-sentence, karpukhin-etal-2020-dense, ni-etal-2022-large}. 
With dual-encoders, each query and document is encoded into a low-dimensional feature vector respectively, and their relevance is measured by a non-parametric similarity function between the embedding vectors \cite{mussmann2016learning}. 
Due to the limited expressivity from the similarity function, dual encoder models often generalize poorly to new tasks with scarce training data \cite{thakur2021beir}. Previous studies use techniques such as data augmentation \cite{wang-etal-2022-gpl,yu2022generate,izacard2022unsupervised, gao-callan-2022-unsupervised, lin2023train, dai2023promptagator}, continual pre-training \cite{chang2020pre, sachan-etal-2021-end, oguz-etal-2022-domain}, task-aware training \cite{xin-etal-2022-zero,cheng-etal-2023-task}, hybrid sparse-dense retrieval \cite{luan-etal-2021-sparse,chen-etal-2022-salient}, or mixed strategy retrieval \cite{ma-etal-2022-open-domain,ma-etal-2023-chain} and so on to improve cross-task generalization performance of dense retrievers.

The motivation of our work echoes in part with multi-vector retrieval, e.g. ColBERT \cite{khattab2020colbert}, DensePhrase \cite{lee-etal-2021-learning, lee-etal-2021-phrase}, ME-BERT \cite{luan-etal-2021-sparse}, and MVR \cite{zhang-etal-2022-multi}, where the retrieval model learns to encode a candidate retrieval unit into multiple vectors to increase model expressivity and improve retrieval granularity \cite{seo-etal-2019-real, humeau2019poly}. Our work instead focuses on the setting where we do not update the dense retriever model or its parameters. 
We show that indexing the retrieval corpus by different granularity can be a simple and orthogonal strategy for improving the generalization of dense retrievers at inference time.

In line with generating retrieval units from the original corpus, \citet{sarthi2024raptor} propose using generative summaries as additional retrieval units alongside the original text, enhancing queries with document-level understanding. In contrast, our work generates propositions to improve queries related to long-tailed entities. These approaches are complementary, as they address different aspects of retrieval enhancement.

The use of propositions as a unit of text representation dates back to the Pyramid method in summarization evaluation \cite{nenkova-passonneau-2004-evaluating}, where a model-generated summary is evaluated by each proposition. Proposition extraction from text has been a long-standing task, with earlier formulations focusing on a structured representation of propositions \cite{etzioni2008open, gildea-jurafsky-2000-automatic}. More recent studies have found success in extracting free-text propositions via few-shot prompting with LLMs \cite{min2023factscore,kamoi2023wice}, or fine-tuning compact-sized models \cite{chen2023subsentence}. 

\textit{Retrieve-then-read}, or more broadly retrieval augmented generation, has recently emerged as a popular paradigm for open-domain question answering \cite{lewis2021retrievalaugmented,jiang2023active,asai2023selfrag}. While earlier works provide up to the top 100 retrieved passages for the downstream reader \cite{izacard-grave-2021-leveraging,kedia-etal-2022-fie}, the amount of context allowed is significantly reduced when using recent large language models \cite{touvron2023llama,yu2023chain}, due to the limited context window length and inability to reason over long context \cite{liu2023lost}. Recent efforts try to improve the quality of the reader context by filtering or compressing the retrieved documents \cite{wang2023learning,xu2023recomp}. Our work offers a new perspective by changing the retrieval granularity, in order to achiev greater information density with a fixed context length.

\section{Conclusion}

This paper studies how the choice of granularity for indexing a corpus, as well as the granularity used in the prompts, influences retrieval and downstream QA performance. Our results show that retrieval by propositions outperforms passage-level and sentence-level retrieval on passage retrieval and downstream QA across five open-domain QA datasets. Our analysis shows that indexing a corpus with finer-grained units enhances the cross-task generalization of dense retrievers and increases the density of question-related information in the prompts. We hope that \datasetname~and our findings will facilitate future research on information retrieval and retrieval-augmented generation.

\section*{Limitations}
The scope of our current study on the granularity of retrieval corpus has the following limitations. (1) \textit{Retrieval Corpus} -- Our study only focuses on Wikipedia as the retrieval corpus, due to the fact that most open-domain QA datasets adopt Wikipedia as the retrieval corpus. (2) \textit{Types of dense retrievers evaluated} -- In the current version of the paper, we evaluate 6 types of popular dense retrievers, most of which follow the bi- or dual-encoder architecture. In future versions, we will include and discuss results on a broader range of dense retrievers. (3) \textit{Language} -- Our current study is limited to English Wikipedia only. We leave the exploration on other languages to future work.

\section*{Ethical Considerations}
This article follows the ACL Code of Ethics. Our work is a foundational research on information retrieval. To the best of our knowledge, we do not find obvious risks related to malicious harmful effects, environmental impact, fairness considerations, or privacy considerations.

\section*{Acknowledgements}
The authors sincerely appreciate anonymous reviewers for helpful discussions and comments. The authors would like to thank Xuanyu Ben Zhou, Ruixin Hong, Ning Dai, and Linfeng Shen for valuable feedback on the project. Xinran Zhao is supported by the ONR Award N000142312840.

\nocite{newman2023controllable}
\bibliography{anthology,custom}

\newpage
\appendix




\newpage

\section{Retrieval Corpus Processing}\label{sec:corpus-details}
The English Wikipedia dump used in this study, released by \citealp{bohnet2022attributed}, was selected because it has been filtered to remove figures, tables, and lists, and is organized into paragraphs. The dump dates back to October 13, 2021.
We have segmented Wikipedia into three retrieval units for this study: 100-word passage chunks, sentences, and propositions.
Paragraphs are divided into 100-word passage chunks using a greedy method. We divide only at the end of sentences to ensure each passage chunk contains complete sentences. As we process the paragraph, we add sentences one by one. If including the next sentence causes the passage chunk to exceed 100 words, we start a new passage chunk with that sentence. However, if the final passage chunk is shorter than 50 words, we merge it with the previous one to avoid overly small segments.
Each passage is further segmented into sentences using the widely used Python SpaCy \footnote{\url{https://spacy.io/}} \longtt{en\_core\_web\_lg} model.
Additionally, each passage is decomposed into propositions by our \textit{Propositionizer} model.
Decomposing the entire Wikipedia corpus requires approximately 500 GPU hours on NVIDIA P100 GPUs using the default implementation in the transformers\footnote{\url{https://huggingface.co/docs/transformers/en/index}} package.
We decomposed 6 million pages into 41 million passages, 114 million sentences, and 257 million propositions. On average, a passage contains 6.3 propositions, and a sentence contains 2.3 propositions.



\section{Training the Propositionizer}
We generated a list of propositions from a given paragraph using GPT-4 with a prompt, as shown in Figure \ref{fig:prompt}. After filtering, 42,857 pairs were used to fine-tune a Flan-T5-Large model. We named the model Propositionizer. The AdamW optimizer was used with a batch size of 64, learning rate of 1e-4, weight decay of 1e-4, and 3 epochs.

To compare the proposition generation performance of different models, we set up a development set and an evaluation metric. The development set contains an additional 1,000 pairs collected by GPT-4 using the same approach as the training set. We evaluated the quality of the predicted propositions by the F1 score of two sets of propositions.
Motivated by the F1 score of two sets of tokens in BertScore, we designed the F1 score for two sets of propositions.
Let $P=\{p_1, ..., p_n\}$ denote the set of labeled propositions and $\hat{P}=\{\hat{p}_1, ..., \hat{p}m\}$ the set of predicted propositions. We use $\mathrm{sim}(p_i, \hat{p}j)$ to represent the similarity between two propositions.
Theoretically, any text similarity metric can be used.
We chose BertScore~\citep{Zhang2020BERTScore} with roberta-large~\cite{liu2020roberta} configuration as our $\mathrm{sim}$ function since we wanted our metric to reflect the semantic difference between propositions.
We define
\begin{equation*}
\begin{aligned}
\mathrm{Recall} &= \frac{1}{|P|} \sum_{p_i \in P} \max_{\hat{p}_j \in \hat{P}} \mathrm{sim}(p_i, \hat{p}_j) \\
\mathrm{Precision} &= \frac{1}{|\hat{P}|} \sum_{\hat{p}_j \in \hat{P}} \max_{p_i \in P} \mathrm{sim}(p_i, \hat{p}_j) \\
\mathrm{F1} &= 2 \cdot \frac{\mathrm{Precision} \cdot \mathrm{Recall}}{\mathrm{Precision} + \mathrm{Recall}}
\end{aligned}
\end{equation*}
Here is a figurative explanation of the F1 score: $\mathrm{Recall}$ represents the percentage of propositions in the labeled set that are similar to those in the generated set, $\mathrm{Precision}$ represents the percentage of propositions in the generated set that are similar to the labeled set, and $\mathrm{F1}$ is the harmonic mean of $\mathrm{Recall}$ and $\mathrm{Precision}$.
$\mathrm{F1}$ is 1 if the two sets are exactly the same, and 0 if any two propositions are semantically different.

\begin{figure}[h]
    \centering
    \includegraphics[width=0.8\columnwidth]{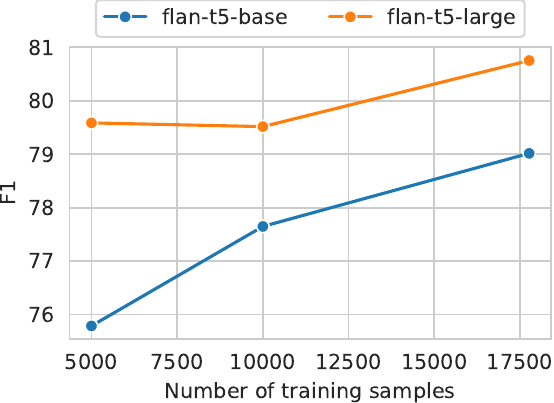}
    \caption{Performance of proposition-level decomposition by models with different sizes and number of training data.}
    \label{fig:prop-f1}
\end{figure}

We conducted a comparative analysis of base-size and large-size Flan-T5 models, which were trained using varying amounts of data (shown in \autoref{fig:prop-f1}). Our findings suggest that larger models, coupled with extensive training data, yield better results. 
The \textit{Propositionizer} presented in this paper attained an F1 score of 0.822. Upon manually reviewing the generated propositions, we found them to be satisfactory.

\section{Quality Analysis of Generated Propositions}
We collected propositions generated from 50 randomly selected passages. There are 408 and 445 propositions generated by GPT-4 and Propositionizer, respectively. The propositions and passages were provided to an expert without knowing which model generated each proposition. The expert annotated three scores from different perspectives for each proposition: (1) whether the proposition is fully supported by the passage, (2) whether the proposition is minimal and cannot be further split into separate propositions, and (3) whether the proposition is self-contained. The scores range from 1 to 3, where 1 means "no," 2 means "maybe," and 3 means "yes." We report the number of cases where the annotation was "no." The detailed instructions are provided in \autoref{tab:quality-anno}.

\section{Offline Indexing}
We used the \texttt{pyserini} and \texttt{faiss} packages to encode retrieval units into embeddings. We exploited multiple GPUs to encode each text unit in groups of 1M units with a batch size of 64. After preprocessing the embeddings, we used an exact search for the inner product (\texttt{faiss.IndexFlatIP}) in all experiments. The plain index of \datasetname is approximately 768GB in size. To reduce memory pressure, the embeddings are split into 8 shards. An approximate nearest neighbor search is conducted per shard before aggregating all results. 

Although the number of propositions is six times that of passages, using efficient indexing techniques can enable sub-linear search times relative to the total count of vectors. Moreover, utilizing GPU parallelism and distributed indexes significantly decreases the online search time. As a result, with proper implementation, we can make proposition retrieval a practically viable and efficient option.

\section{Retriever Models and QA Models}
We used \longtt{transformers} and \longtt{sentence-transformers} packages for the model implementation. 
We used the following checkpoints released on HuggingFace: 
SimCSE~(\longtt{princeton-nlp/unsup-simcse-bert-base-uncased}), 
Contriever~(\longtt{facebook/contriever}), 
DPR~(\longtt{facebook/dpr-ctx\_encoder-multiset-base}, \longtt{facebook/dpr-question\_encoder-multiset-base}), 
GTR~(\longtt{sentence-transformers/gtr-t5-base}).

We use T5-large size Fusion-in-decoder model (\longtt{nq\_reader\_large}) released by the authors in \url{https://github.com/facebookresearch/FiD}. We use HuggingFace checkpoint (\longtt{meta-llama/Llama-2-7b}) for LLaMA-2-7B.

\section{Additional Results}

\begin{figure*}[h]
    \centering\small
    \includegraphics[width=0.75\textwidth]{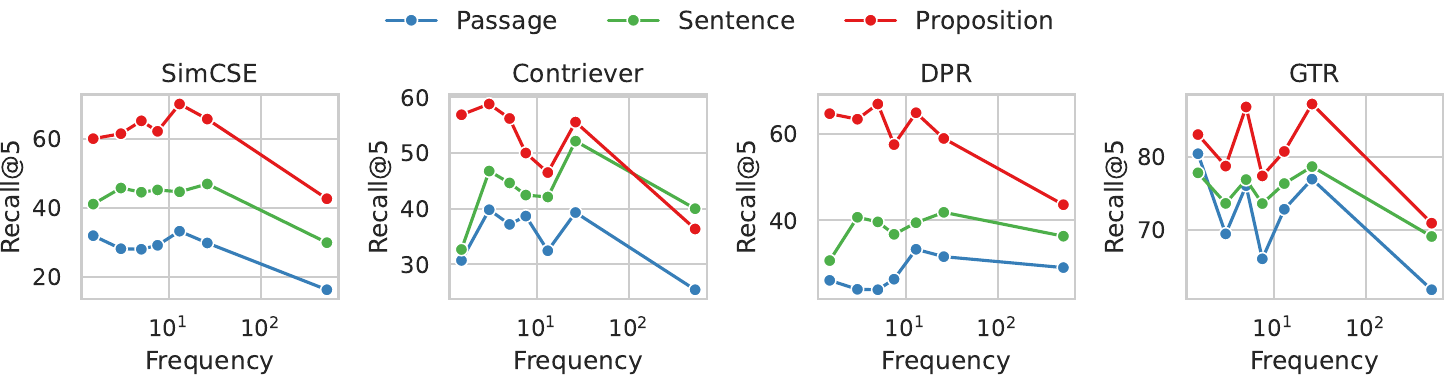} \\
    (a) Where was [X] born?
    
    \includegraphics[width=0.75\textwidth]{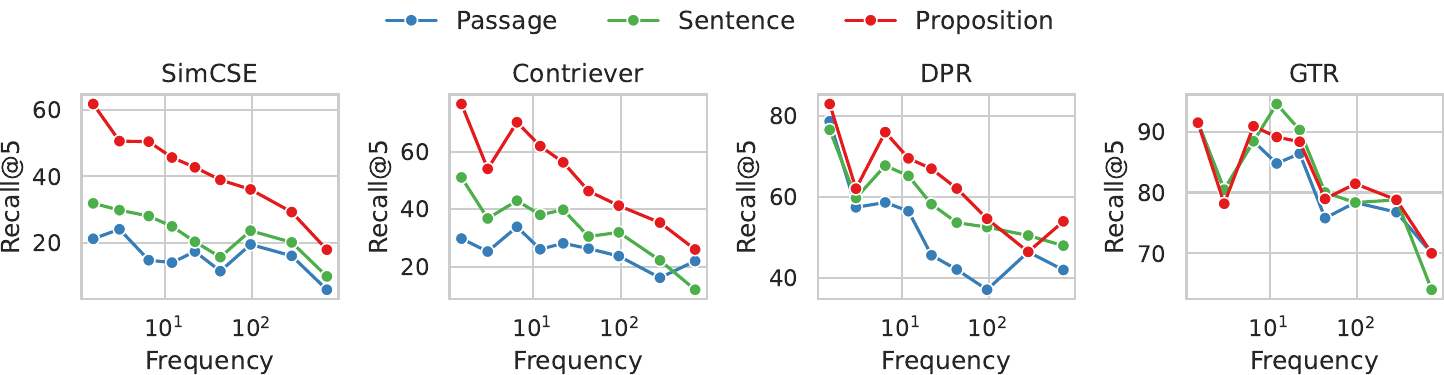}  \\
    (b) Who was [X] created by?
    \caption{Document retrieval recall vs. the frequency of the target entity in each question from the \textit{Entity Questions} dataset. We display the performance of two relations. }
    \label{fig:entity-all}
\end{figure*}

In Section \ref{ssec:generalization}, we demonstrated the advantage of retrieval by proposition over retrieval by sentence, particularly as the population of the entity decreases in EQ. We used the occurrence in the top-1000 paragraphs retrieved by BM25 as a proxy for frequency, rather than counting the number of hyperlinks to the entity used in \citealp{sciavolino2021simple}. Therefore, the trend in the performance versus frequency plot shows some differences (\autoref{fig:entity-all}) between our results and those in \citealp{sciavolino2021simple}. For example, some entities are ambiguous (e.g., \textit{1992}, a TV series). In such cases, the occurrence of the surface form of the entity is large. Simultaneously, questions related to ambiguous entities are challenging to answer, leading to lower recall. 

\begin{figure*}[h]
    \centering
    \includegraphics[width=0.99\textwidth]{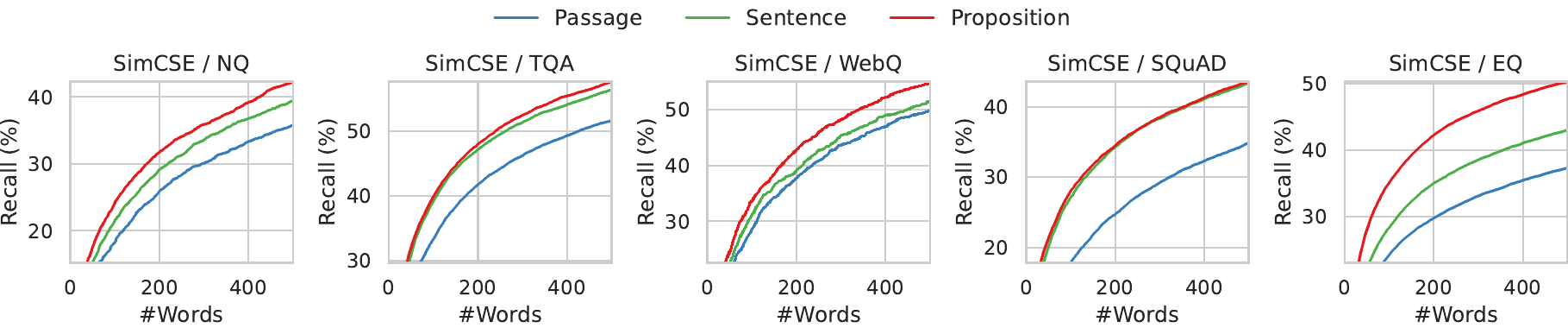}
    \includegraphics[width=0.99\textwidth]{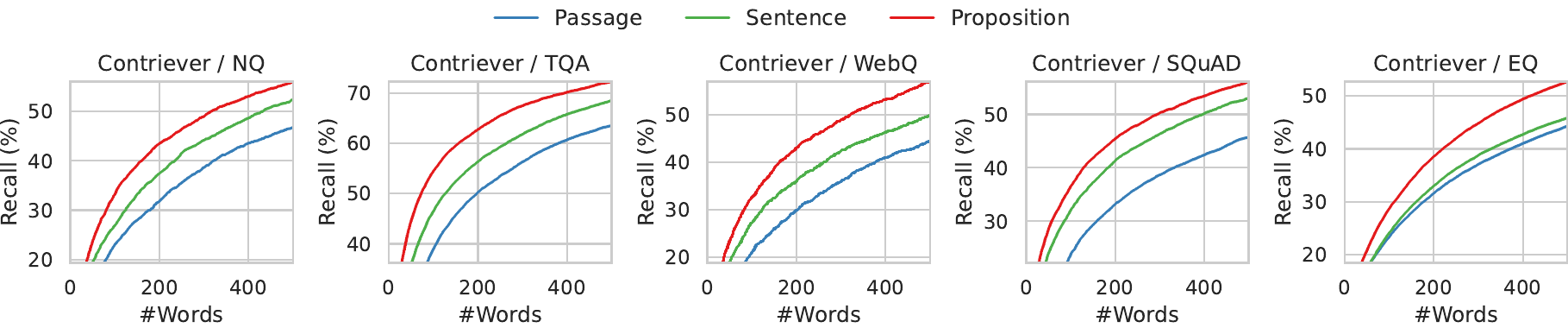}
    \includegraphics[width=0.99\textwidth]{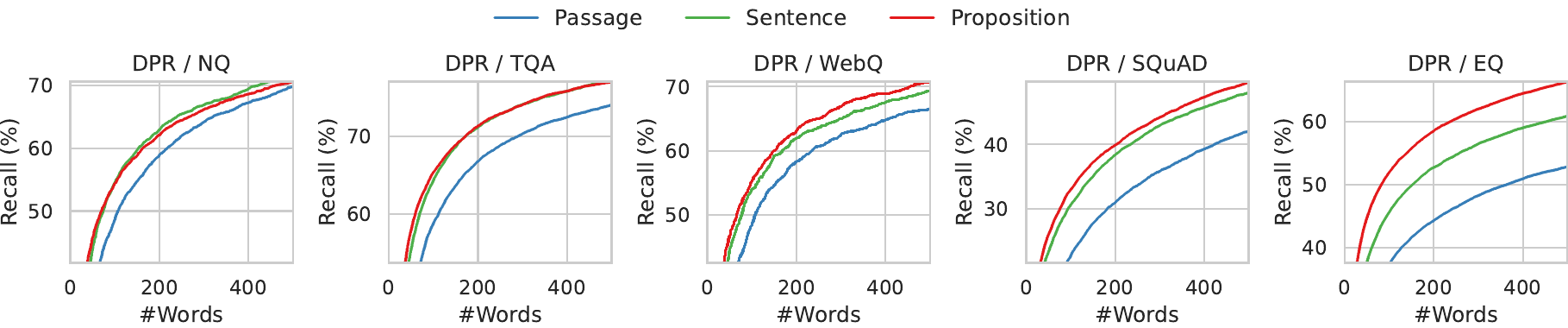}
    \includegraphics[width=0.99\textwidth]{figures/position_gtr.pdf}
    \vspace{-0.1in}
    \caption{Recall of the gold answer in the retrieved text limited to first $k$ words. Finer-grained retrieval has a higher recall across all numbers of words.}
    \label{fig:position-all}
\end{figure*}

In Section \ref{sec:density}, we discussed the recall of answers in the retrieved text with respect to the context length. We further illustrate the performance trends of six dense retrievers, as detailed in \autoref{fig:position-all}.
The results indicate that the recall rate of propositions consistently outperforms that of sentences and passages.
Our findings lead to the conclusion that question-related density is greater in proposition units compared to sentences and passages.

\section{Error Case Study}

To understand the source of errors from each type of retrieval granularity, we present and discuss four typical examples of mistakes in     \autoref{tab:case} and \autoref{tab:case-app}. With each example, we show the question and its corresponding top-1 retrieved text unit by the GTR retriever across the three granularities.

We observe that with passage-level retrieval, the ambiguity of an entity or its references presents a challenge for dense retrievers, which echoes findings from \cite{min-etal-2020-ambigqa}. 
For instance, in example Q1, the question asks for ``\textit{Super Bowl 50}'', but the retrieved passage and sentence refers to ``\textit{Super Bowl 5}''. 
In Example Q2, passage retrieval fails to identify the part referring to the correct ``\textit{atomic number}''. Instead, the top-1 retrieved passage mentions ``\textit{atomic number}'' in a different and irrelevant context to the question. 
Retrieval by sentences can also have a similar problem as retrieval by passages like Example Q1. Also, retrieval by sentences faces another challenge of lacking context.
In Example Q3 (shown in \autoref{tab:case-app}), sentence-based retrieval fails as the correct sentence in the retrieved passage uses ``\textit{it}'' to refer to the pericardial sac.

Retrieval by propositions tackles the aforementioned problems by ensuring each retrieval unit contains one piece of fact only and necessary context is incorporated in the propositions.
However, proposition-based retrieval faces challenges with questions that involve multi-hop reasoning over long-range textual analysis.
In Example Q4 (shown in \autoref{tab:case-app}), the retrieved passage separately describes the actor's name and the character they portray. There is not a single proposition that entails both the question and the answer. 

\begin{table*}[t]
    \centering
\resizebox{\textwidth}{!}{%
\footnotesize
\begin{tabular}{p{0.35\textwidth} p{0.3\textwidth} p{0.35\textwidth}}
\toprule
\multicolumn{1}{c}{\textbf{Passage Retrieval}} &
\multicolumn{1}{c}{\textbf{Sentence Retrieval}} &
\multicolumn{1}{c}{\textbf{Proposition Retrieval}} 
\\ \midrule
\multicolumn{3}{l}{Q1: What was the theme of Super Bowl 50?}
\\ \midrule
Title: Super Bowl X \hfill \textcolor{red}{\xmark} \newline
The overall theme of the Super Bowl entertainment was to celebrate the United States Bicentennial. Each Cowboys and Steelers player wore a special patch with the Bicentennial logo on their jerseys... 
&
Title: Super Bowl X \hfill \textcolor{red}{\xmark}    \newline
The overall theme of the Super Bowl entertainment was to celebrate the United States Bicentennial. 
&
Title: Super Bowl XLV \hfill \textcolor{green}{\cmark}  \newline 
\textcolor{lightgray}{... As this was the 50th Super Bowl game, the league }\ul{[Super Bowl 50] emphasized the "golden anniversary"} \textcolor{lightgray}{with various gold-themed initiatives during the 2015 season, as well as...} 
\\ \midrule

\multicolumn{3}{l}{Q2: The atomic number of indium which belongs to 5th period is?}  
\\ \midrule
Title: Period 5 element \hfill \textcolor{red}{\xmark} \newline
The periodic table is laid out in rows to illustrate recurring (periodic) trends in the chemical behaviour of the elements as their atomic number increases: ... 
&
 Title: Period 5 element \hfill \textcolor{green}{\cmark} \newline
Indium is a chemical element with the symbol In and \uline{atomic number 49}. 
&
Title: Period 5 element \hfill \textcolor{green}{\cmark}  \newline
\textcolor{lightgray}{Indium is a chemical element with the symbol In and} [Indium has a] \uline{atomic number 49}. \textcolor{lightgray}{This rare, very soft, malleable ...} 

\\ \bottomrule
\end{tabular}%
}
\caption{
Example cases where top-1 retrieved text unit of each retrieval granularity fails to provide the correct answer.
The \underline{underlined text} is the correct answer. 
The \textcolor{lightgray}{gray text} is the context of propositions, but it is for illustration purpose only and not provided to the retrievers and downstream QA models.
}
\vspace{-5pt}
\label{tab:case}
\end{table*}

\begin{table*}[t]
    \centering
\resizebox{\textwidth}{!}{%
\footnotesize
\begin{tabular}{p{0.35\textwidth} p{0.3\textwidth} p{0.35\textwidth}}
\toprule
\multicolumn{1}{c}{\textbf{Passage Retrieval}} &
\multicolumn{1}{c}{\textbf{Sentence Retrieval}} &
\multicolumn{1}{c}{\textbf{Proposition Retrieval}} 
\\ \midrule

\multicolumn{3}{l}{Q3: What is the function of the pericardial sac?}  \\ \midrule
Title: Pericardium \hfill \textcolor{green}{\cmark} \newline
The pericardium, also called pericardial sac ... It separates the heart from interference of other structures, protects \uline{it against infection and blunt trauma, and lubricates the heart's movements.} 
&
Title: Pericardium \hfill \textcolor{red}{\xmark} \newline
The pericardium, also called pericardial sac, is a double-walled sac containing the heart and the roots of the great vessels. 
&
Title: Cardiac muscle \hfill \textcolor{green}{\cmark}  \newline
\textcolor{lightgray}{On the outer aspect of the myocardium is the epicardium which forms part of} \ul{the pericardial sac that surrounds, protects, and lubricates the heart.} 
\\ \midrule

\multicolumn{3}{l}{Q4: What is the main character's name in layer cake?}  \\ \midrule

Title: Layer Cake (film) \hfill \textcolor{green}{\cmark}  \newline
... The film's plot revolves around a London-based criminal, played by Daniel Craig, ... Craig's character is \underline{unnamed} in the film and is listed in the credits as \underline{"XXXX"}.
&
Title: Angelic Layer \hfill \textcolor{red}{\xmark} \newline
The primary protagonist is Misaki Suzuhara.
&
Title: Plot twist \hfill \textcolor{red}{\xmark} \newline
\textcolor{lightgray}{Sometimes the audience may discover that} the true identity of a character is \textcolor{lightgray}{, in fact,} unknown [in Layer Cake] \textcolor{lightgray}{, as in Layer Cake or the eponymous assassins in V for Vendetta and The Day of the Jackal.}
\\ \bottomrule
\end{tabular}%
}
\caption{
Example cases where top-1 retrieved text unit of each retrieval granularity fails to provide the correct answer.
The \underline{underlined text} is the correct answer. 
The \textcolor{lightgray}{gray text} is the context of propositions, but it is for illustration purpose only and not provided to the retrievers and downstream QA models.
}
\vspace{-5pt}
\label{tab:case-app}
\end{table*}

\begin{table*}[]
    \centering
    \begin{tabular}{p{0.9\textwidth}}
\toprule
        \textbf{Is the proposition fully supported by the passage?}   
\textbf{No} : The information provided relates to the proposition, but there are some gaps or inconsistencies that prevent full support.
\textbf{Maybe} : The information provided supports the proposition adequately, covering most aspects well; however, minor details or implications might not be fully explored or clarified.
\textbf{Yes}: The information provided clearly and comprehensively addresses all aspects of the proposition, leaving no relevant details unexplained or ambiguous.\\ \hline
       \textbf{Should the given propositions be further split into separate propositions?  }
\textbf{No}: The proposition has a compound structure that could be separated into distinct propositions.
\textbf{Maybe}: The proposition is mostly straightforward with a single main idea and perhaps a minor additional detail. Splitting might enhance clarity but is not strictly necessary.
\textbf{Yes}: The proposition is already concise and does not contain a compound structure. Splitting it into separate propositions would likely reduce clarity.
 \\\hline
   \textbf{Is the given proposition self-contained? }
\textbf{No}: The proposition contains pronouns, terms, or references whose full names or meanings are not in the proposition.
\textbf{Maybe}: The proposition is almost entirely self-contained, with only a few minor terms that might be ambiguous without additional context.
\textbf{Yes}: The proposition is a self-contained claim without any ambiguities, fully understandable on its own.
\\ \bottomrule
    \end{tabular}
    \caption{Instructions for data annotation in analyzing the quality of generated propositions.}
    \label{tab:quality-anno}
\end{table*}

\begin{figure*}[t]
\begin{prompt}{Passage $\Rightarrow$ Propositions}
Decompose the "Content" into clear and simple propositions, ensuring they are interpretable out of context. 
\begin{enumerate}[leftmargin=*, itemsep=0em]
    \item Split compound sentence into simple sentences. Maintain the original phrasing from the input whenever possible. 
    \item For any named entity that is accompanied by additional descriptive information, separate this information into its own distinct proposition. 
    \item Decontextualize the proposition by adding necessary modifier to nouns or entire sentences and replacing pronouns (e.g., "it", "he", "she", "they", "this", "that") with the full name of the entities they refer to. 
    \item Present the results as a list of strings, formatted in JSON.
\end{enumerate}

\textbf{Input}: 
Title:  Ēostre. Section: Theories and interpretations, Connection to Easter Hares. Content: The earliest evidence for the Easter Hare (Osterhase) was recorded in south-west Germany in 1678 by the professor of medicine Georg Franck von Franckenau, but it remained unknown in other parts of Germany until the 18th century. Scholar Richard Sermon writes that "hares were frequently seen in gardens in spring, and thus may have served as a convenient explanation for the origin of the colored eggs hidden there for children. Alternatively, there is a European tradition that hares laid eggs, since a hare's scratch or form and a lapwing's nest look very similar, and both occur on grassland and are first seen in the spring. In the nineteenth century the influence of Easter cards, toys, and books was to make the Easter Hare/Rabbit popular throughout Europe. German immigrants then exported the custom to Britain and America where it evolved into the Easter Bunny."

\textbf{Output}:
[
  "The earliest evidence for the Easter Hare was recorded in south-west Germany in 1678 by Georg Franck von Franckenau.",
  "Georg Franck von Franckenau was a professor of medicine.",
  "The evidence for the Easter Hare remained unknown in other parts of Germany until the 18th century.",
  "Richard Sermon was a scholar.",
  "Richard Sermon writes a hypothesis about the possible explanation for the connection between hares and the tradition during Easter",
  "Hares were frequently seen in gardens in spring.",
  "Hares may have served as a convenient explanation for the origin of the colored eggs hidden in gardens for children.",
  "There is a European tradition that hares laid eggs.",
  "A hare's scratch or form and a lapwing's nest look very similar.",
  "Both hares and lapwing's nests occur on grassland and are first seen in the spring.",
  "In the nineteenth century the influence of Easter cards, toys, and books was to make the Easter Hare/Rabbit popular throughout Europe.",
  "German immigrants exported the custom of the Easter Hare/Rabbit to Britain and America.",
  "The custom of the Easter Hare/Rabbit evolved into the Easter Bunny in Britain and America."
]
\\

\textbf{Input}: <\textit{a new passage}>

\textbf{Output}:
\end{prompt}
\caption{Prompt for generating propositions from a passage using GPT-4.}\label{fig:prompt}
\end{figure*}

\begin{figure*}[t]
\begin{prompt}{Open-domain QA for LLaMA-2-7B}
... [demonstrations] ...
\\

\textbf{Refer to the passages below and answer the following question with just a few words.}

\textbf{Title}: 1972 in spaceflight. \textbf{Passage}: In 1972, humanity's last crewed mission to the Moon of the 20th century was Apollo 17.

\textbf{Title}: 1970s. \textbf{Passage}: Apollo 17 Astronaut Gene Cernan becomes the last man on the Moon on December 13, 1972.

\textbf{Title}: List of Apollo missions

\textbf{Refer to the context above and answer the following question with just a few words.}

Question: when was the last time anyone was on the moon

The answer is

\end{prompt}
\caption{Prompt for retrieval-augmented generation of open-domain QA for the LLaMA-2-7B model.}\label{fig:llama-prompt}
\end{figure*}

\end{document}